\pdfoutput=1
\documentclass[11pt]{article}

\usepackage[]{EMNLP2022}

\usepackage{times}
\usepackage{latexsym}
\usepackage{color}
\usepackage{xcolor}
\usepackage{colortbl}
\definecolor{Gray}{gray}{0.9}
\definecolor{LightGray}{gray}{0.94}

\usepackage[T1]{fontenc}
\usepackage[utf8]{inputenc}

\usepackage{microtype}

\usepackage{graphicx}
\usepackage{caption}
\usepackage{algorithm}
\usepackage{algorithmic}
\usepackage{subcaption}
\usepackage{multirow}
\usepackage{amsmath}
\usepackage{bbm}
\usepackage{natbib}
\usepackage{hyperref}
\usepackage[normalem]{ulem}

\newcommand\raymond[1]{{\leavevmode\color{black}#1}}

\newcommand\change[1]{{\leavevmode\color{red}#1}}
\newcommand{\cut}[1]{\textcolor{gray}{\sout{#1}}}

\renewcommand{\change}[1]{{\leavevmode\color{black}#1}} 
\renewcommand{\cut}[1]{\unskip}

\title{
Human 
Guided
Exploitation of 
Interpretable Attention Patterns in 
Summarization and Topic Segmentation}

\author{Raymond Li\footnotemark[2]~, Wen Xiao\footnotemark[2]~, Linzi Xing\footnotemark[2]~, Lanjun Wang\footnotemark[3]~\hspace{.1em}\thanks{~~Corresponding author.}~, Gabriel Murray\footnotemark[4]~, Giuseppe Carenini\footnotemark[2]\\
\footnotemark[2]\hspace{.2em} University of British Columbia, Vancouver, BC, Canada \\
\footnotemark[3]\hspace{.2em} Tianjin University, Tianjin, China \\
\footnotemark[4]\hspace{.2em} University of Fraser Valley, Abbotsford, BC, Canada \\
\texttt{\{\href{mailto:raymondl@cs.ubc.ca}{\texttt{raymondl}}, \href{mailto:xiaowen3@cs.ubc.ca}{\texttt{xiaowen3}}, \href{mailto:lzxing@cs.ubc.ca}{\texttt{lzxing}}, \href{mailto:carenini@cs.ubc.ca}{\texttt{carenini}}\}@cs.ubc.ca} \\
\href{mailto:wanglanjun@tju.edu.cn}{\texttt{wanglanjun@tju.edu.cn}}\hspace{1em}  \href{mailto:gabriel.murray@ufv.ca}{\texttt{gabriel.murray@ufv.ca}}
}

\begin{document}
\maketitle
\begin{abstract}
The 
\cut{transformer multi-head self-attention mechanism}
\change{multi-head self-attention mechanism of the transformer model}
has been thoroughly investigated recently. 
% On one hand
In one vein of study, researchers are interested in understanding why and how transformers work. 
% On the other hand
In another vein, researchers propose new attention augmentation methods to make transformers more accurate, efficient and interpretable. In this paper, we 
% synergize
combine these two lines of research in a human-in-the-loop pipeline to first discover 
% find 
important task-specific attention patterns. 
% Then those patterns are applied, not only to the original model, but also to smaller models 
Then those patterns are injected,
% applied
 not only to smaller models, but also to the original model.
% , as a human-guided knowledge distillation process.
% The benefits of our pipeline are demonstrated in a case study with the extractive summarization task.  After finding three meaningful attention patterns in the popular BERTSum model, experiments indicate that when we inject such patterns, both the original and the smaller model show improvements in performance and arguably interpretability. 
The benefits of our pipeline and discovered
% found
patterns are demonstrated in 
% our
two case studies with extractive summarization and topic segmentation. After discovering
% finding 
interpretable patterns in BERT-based models fine-tuned for the two downstream tasks,
% popular BERTSum model,
experiments indicate that when we inject the patterns into attention heads, the models show considerable improvements in accuracy and efficiency. 
% \change{Our code base is available at \url{https://github.com/raymondzmc/Attention-Pattern-Exploitation}}
\end{abstract}

% Abstract for copying to text-box
% The transformer multi-head self-attention mechanism has been thoroughly investigated recently. In one vein of study, researchers are interested in understanding why and how transformers work. In another vein, researchers propose new attention augmentation methods to make transformers more accurate, efficient and interpretable. In this paper, we combine these two lines of research in a human-in-the-loop pipeline to first discover important task-specific attention patterns. Then those patterns are injected, not only to smaller models, but also to the original model. The benefits of our pipeline and found patterns are demonstrated in two case studies with extractive summarization and topic segmentation. After discovering interpretable patterns in BERT-based models fine-tuned for the two downstream tasks, experiments indicate that when we inject the patterns into attention heads, the models show considerable improvements in accuracy and efficiency.

\section{Introduction}
With transformer-based models \citep{transformer} dominating the leaderboard for many key NLP tasks \cut{like} \change{such as} summarization \citep{liu-lapata-2019-text}, topic segmentation \citep{lukasik-etal-2020-text}, and sentiment analysis \cite{sentiment_bert}, their core multi-head self-attention mechanism has also been thoroughly investigated. In particular, to explain why and how transformers work, researchers have analyzed the learnt self-attention matrices of 
% language models or task-specific models
trained transformer models
(e.g., \citet{raganato-tiedemann-2018-analysis, kovaleva-etal-2019-revealing}),
% with \citet{voita-etal-2019-analyzing} for instance, exploring the patterns of attention heads in neural machine translation during pruning. 
with \citet{vig-belinkov-2019-analyzing} for instance, exploring the attention patterns in BERT \citep{devlin-etal-2019-bert} and GPT-2 \cite{radford2019language}, as well as analyzing their alignment with syntax.

Meanwhile, a parallel line of research has explored
% with
injecting predefined patterns into attention matrices of transformers in an attempt 
% reducing
to reduce the run-time complexity of self-attention while maintaining competitive accuracy.
% performance.
This can be done by either replacing the
% with learnt 
attention weights with a fixed matrix \citep{raganato-etal-2020-fixed, tay2020synthesizer, xiao-etal-2020-really}; or alternatively by guiding the attention weights through more flexible masking strategies \citep{mihaylov-frank-2019-discourse, child2019generating, guo-etal-2019-star, li2019enhancing, Beltagy2020Longformer, zaheer2020big, bai-etal-2021-syntax}. 
% In addition, we found that our strategy can also be applied to pre-trained transformer models by injecting the patterns into newly added attention heads using techniques such as Projected Attention Layers (PAL) \citep{stickland2019bert}.

In this work, we propose and test a novel human-in-the-loop pipeline that to the best of our knowledge is the first attempt 
\cut{trying}
to combine research on analyzing self-attention with work on injecting patterns into attention matrices. 
To start, human users visually explore the attention matrices of transformers to identify task-specific patterns that could be formalized as a predicate. After quantitatively evaluating the patterns on the validation set, they can be injected into attention heads of transformer models to \change{simultaneously} improve \cut{both} task accuracy and \cut{efficiency} \change{make the model more efficient by sparsifying the attention matrices}\footnote{The implementation of our work is publicly available at: \href{https://github.com/raymondzmc/Attention-Pattern-Exploitation}{https://github.com/raymondzmc/Attention-Pattern-Exploitation}}. This is in contrast to previous work that mostly focuses on the trade-off between two metrics.

In both scenarios, we argue the interpretability of the resulting model is improved. We provide a justification of our claim based on the Predictive, Descriptive, and Relevant (PDR) framework proposed by \citet{murdoch2019interpretable}. Specifically, by injecting human-interpretable patterns into the model, we increase the model's descriptive accuracy by explicitly encoding useful relationships between input tokens in the attention weights while simultaneously improving the predictive accuracy in task performance. Further, the patterns are relevant for the problem since they are discovered in the human-in-the-loop process and are verified to be important for the task.
% Alternative summary
% In both scenarios, we argue the interpretability of the resulting model is improved. We provide a discussion of our interpretability claim based on the Predictive, Descriptive, and Relevant (PDR) framework proposed by \citet{murdoch2019interpretable}.

In order to test the feasibility and potential benefits of our approach, we run two case studies on the \raymond{tasks of extractive summarization and topic segmentation} using BERT-based models, 
% in particular using the popular BERTSum model \cite{liu-lapata-2019-text}, 
and we find
that: 
\textit{(i)} 
% For some of the important heads, the  patterns they learn do have interpretable meaning, 
Some of the important heads do have patterns with interpretable meaning,
either lexical, local or positional. For instance, the 
% match token
matching token (i.e. the trend of attending to other tokens with the same id) is an important clue for the summarization model. 
% \cut{\textit{(ii)} By applying the patterns back into the original model through PAL, the resulting model can achieve better task performance and stronger interpretability. \textit{(iii)} Remarkably, the human-guided knowledge distilled model performs much better than the vanilla transformer (baseline), and can be competitive even with fine-tuned BERT distilled models, which by distilling all aspects of the model have an a priori substantial advantage over our technique that only distils attention patterns.} 
\raymond{\textit{(ii)} 
% The extracted patterns can be injected by enforcing the corresponding predicate as a constraint through masking and fixing the attention weights
We show that when the discovered
% extracted
patterns are injected 
% applied
to the attention heads of transformer models, both the task accuracy and efficiency of the model can be significantly improved. \textit{(iii)} Additionally, we also propose a strategy to improve the performance of pretrained transformer models by injecting patterns through PALs.
% in both cross-dataset and cross-task settings
}
% in both intra-domain and inter-domain settings.

\section{Related Work}

\change{
In \S\ref{subsec:attanalysis} and \S\ref{subsec:attaug}, we describe the two lines of research that our work aims to combine.
% synergize. 
% In \S\ref{subsec:extsumm} and \S\ref{subsec:topicseg}
% In \S\ref{subsec:related-work-models}, we describe the two NLP tasks used in our case studies. Finally, we summarize the recent trends on enhancing the interpretability of NLP neural models in
% trend for the interpretability of NLP models in
% \S\ref{subsec:interpretability}.
\S\ref{subsec:interpretability} summarizes %the
recent trends on enhancing the interpretability of neural NLP models, while  \S\ref{subsec:related-work-models} introduces the two NLP tasks used for our case studies.
}

\subsection{Attention Analysis in Transformers}
\label{subsec:attanalysis}

Various works have investigated the attention head matrices in transformers \citep{raganato-tiedemann-2018-analysis, clark-etal-2019-bert,kovaleva-etal-2019-revealing,zhao-bethard-2020-berts, xiao-etal-2021-predicting}, often with the aid of visualization tools \citep{vig-2019-multiscale, hoover-etal-2020-exbert, li-etal-2021-t3}. For examples, \citet{vig-belinkov-2019-analyzing} visually explore attention patterns in BERT and GPT-2, analyzing their alignment with syntax. While \citet{voita-etal-2019-analyzing} characterize the functions of the attention heads in Machine Translation (MT)
% MT
models (positional, syntactic, \change{and} rare words), and evaluate the importance of those head functions. More recently, \citet{bian-etal-2021-attention} find the redundancy in BERT's attention patterns to be both phase-independent (pretrained 
% vs 
and fine-tuned) and task-agnostic. Lastly, \citet{huber-carenini-2022-towards} infer discourse structures from the attention patterns of language models 
% (BERT, BART)
(BERT and BART), and find discourse information to be consistently captured in the same heads even when fine-tuned for 
different
% downstream 
tasks.
In this paper, we also aim to find task-specific important attention patterns, but in contrast to previous work that identifies and categorizes attention patterns, we propose a pipeline to leverage these patterns in improving models' performance and interpretability.
% by using the visual interface, T\textsuperscript{3}-Vis~\citep{li-etal-2021-t3}.

\subsection{Attention Augmentation}
\label{subsec:attaug}

We organize the related work on augmenting attention matrices into two categories. 
% The first category completely replaces 
In the first category,
attention weights are completely replaced with a fixed matrix. For example, \citet{raganato-etal-2020-fixed} use fixed positional patterns in MT models and demonstrate benefits for low-resource scenarios, while \citet{tay2020synthesizer} replace the weights computed using dot-product self-attention with a random matrix, and report comparable performance with standard transformers. Later on, \citet{xiao-etal-2020-really} expand their work by using embedded RST-style discourse trees as fixed attention matrices and show the effectiveness of discourse-based attention matrices for extractive summarization. 
In contrast, in the second category of attention augmentation works, masks are applied on top of the attention weights to either inject linguistic information \citep{yang-etal-2018-modeling, mihaylov-frank-2019-discourse} or improve the efficiency of self-attention via fixed patterns \citep{child2019generating, guo-etal-2019-star, li2019enhancing, ainslie-etal-2020-etc}. \change{Just to describe a few prominent examples, \citet{strubell-etal-2018-linguistically} use bi-affine attention to learn syntactic dependencies in attention heads}, and \citet{bai-etal-2021-syntax} inject syntactic structures into BERT through extra attention layers. Concurrently, while \citet{Beltagy2020Longformer} use diagonal/vertical/horizontal patterns to respectively model local and global context, \citet{zaheer2020big} add patterns randomly by drawing inspiration from graph theory.
% By comparison, while in all these previous works the designing of pre-defined patterns requires extensive trial and error, \change{focusing on improving either the accuracy or efficiency at the expense of the other,} in this paper we explore a strategy of discovering 
% % finding
% and assessing important attention patterns interactively (\S\ref{subsec:extract_patterns}). 
In comparison, while in all previous works the designing of pre-defined patterns requires extensive trial and error, and only improve upon either the accuracy or efficiency at the expense of the other, we explore a strategy of discovering 
% % finding
and assessing important attention patterns interactively in this paper.
Not only do the discovered
% extracted
patterns help improve performance in terms of \change{both} accuracy and efficiency, \cut{but} they also reveal valuable insights regarding the internal workings of pretrained language models.
% Rather than designing task-specific patterns through extensive trial and error, we explore the extraction of attention patterns directly from pretrained models based on their importance on the respective task. 
% Not only do the extracted patterns help improve performance in terms of accuracy and efficiency, they also reveal insights regarding the internal workings of these pretrained language models and provide useful clues to develop more accurate and efficient models in the future.

% \subsection{\cut{Knowledge Distillation in NLP}}
% Knowledge distillation (KD) \cite{hinton2015distilling} aims to compress a large teacher model into a smaller student model. This is achieved by training the student network to mimic the behaviors of the teacher model in order to obtain a competitive performance. Specifically, the type of knowledge distilled in KD can be categorized into three types \citep{gou2021knowledge}: response-based (e.g., DistilBert \citep{Sanh2019DistilBERTAD}), feature-based (e.g.,TinyBERT \citep{jiao-etal-2020-tinybert}, MobileBERT \citep{sun-etal-2020-mobilebert}, MiniLM \citep{wang-2020-minilm}), and relation-based (e.g. Multi-head Graph Distillation \citep{lee-2019-mhgd}). Our approach can be considered as of type relation-based because we exploit the relations between tokens in the attention weights. Importantly, the key difference from existing KD works, is that in our approach knowledge is explicitly extracted from the teacher model by a human, and such knowledge becomes a transparent property of the student model, 
% arguably increasing its interpretability.

\subsection{Model Interpretability}
\label{subsec:interpretability}
In the context of Machine Learning, interpretability can be defined as the description of the internals of a model in a way that is understandable to humans \citep{gilpin2018explaining}. 
% This is in contrast  This includes the concept of causality,
% % to the closely-related but distinct concept of causality,
% which explains the changes in output due to a perturbation in the input or model component \citep{doshi2017towards}.
With the rise of deep learning, various techniques have been proposed to interpret the inner workings of neural NLP models. For example, probing classifiers are often used for finding linguistic or knowledge information learned by neural networks \citep{conneau-etal-2018-cram, TenneyXCWPMKDBD19, pimentel-etal-2020-information, voita-titov-2020-information, hou-sachan-2021-birds, aghazadeh-etal-2022-metaphors}, while behaviour testing aims at understanding how models behave through inferences under different controlled settings \citep{mccoy-etal-2019-right, ross-pavlick-2019-well, ribeiro-etal-2020-beyond, koh2021wilds, goel-etal-2021-robustness}. In contrast, our work is %related to 
an example of making interpretability an inherent attribute of the neural models (e.g. \citet{chen-ji-2020-learning, hu-etal-2021-r2d2}), 
% where 
with human-distinguishable patterns revealing insights regarding a subset of parameters in the model.

% \subsection{Extractive Summarization}
% \label{subsec:extsumm}
\subsection{NLP Tasks used in the two Case Studies}
\label{subsec:related-work-models}
Extractive summarization is the task 
% to pick
of picking the most representative sentences as the summary for the given document(s). Current state-of-the-art models, which are mostly based on large-scale pretrained language models \cite{liu-lapata-2019-text,zhong-etal-2020-extractive, jia-etal-2020-neural, ruan-etal-2022-histruct}, can deliver good performance, but why and how such models work so well still remain an open question. In our case study, we adopt 
% the
% run a case study for our proposed pipeline on the extractive summarization task, exploring the discovery, assessment and application of useful attention patterns in the context of the 
% transformer-based 
the popular BERTSum 
% model 
\citep{liu-lapata-2019-text}.

% \subsection{Topic Segmentation}
% \label{subsec:topicseg}
Topic segmentation is the task of breaking stretches of running text into smaller topical-coherent segments consisting of one or more sentences addressing a common topic. 
Recently, more research work frames the task
%problem in 
in the supervised learning paradigm and uses neural models such as Bi-LSTMs \citep{koshorek-etal-2018-text, xing-etal-2020-improving} and transformer \citep{glavas-2020-two, lo-etal-2021-transformer-pre} as the backbone, due to the availability of large-scale labeled benchmarks sampled from \textit{Wikipedia}. These proposed neural topic segmentation models achieve state-of-the-art performance on monologue text by formulating the problem as a sequence labeling task, 
% where a prediction is made for every sentence in the document indicating whether or not it ends a topical segment.
where the predicted label of each sentence indicates whether or not it is the end of a segment.
In our case study, we adopt Cross-Segment BERT \citep{lukasik-etal-2020-text}.

\begin{figure*}[!ht]
    \centering
    \includegraphics[width=\linewidth]{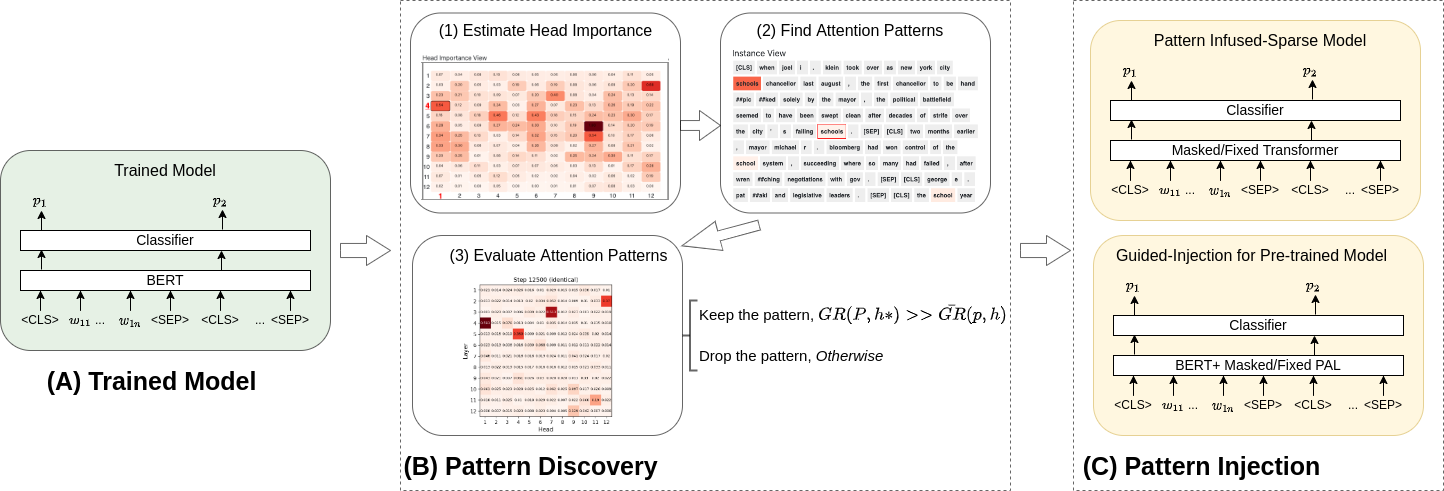}
    \caption{\cut{The diagram of our proposed generic pipeline, which contains two main parts - pattern discovery
    % extraction 
    (middle) and pattern injection 
    % application
    (right), given a trained task-specific model (left)}\change{The overview of our proposed generic pipeline. Given \textbf{(A)} a trained model for a specific task, our pipeline can be divided into two main parts: \textbf{(B)} pattern discovery and \textbf{(C)} pattern injection.} 
    % {\footnotemark}.
    }
    \label{fig:pipeline}
\end{figure*}

\section{Proposed Generic Pipeline}

% In this section, 
As an overview, we first
% we will 
briefly describe the proposed pipeline (\autoref{fig:pipeline}). Specifically, given a trained model, 
users are
% % supposed 
asked to first discover
% % extract 
important patterns using
% %on 
the visual interface~\citep{li-etal-2021-t3} 
% is used to interactively discover important patterns
% (middle part in \autoref{fig:pipeline})
by following three steps: \\
\textit{Step 1} (\S \ref{sec:find_patterns_step1}): Estimate the importance scores for all the heads on the validation set, and find important heads that stand out. 
% (Sec.\ref{sec:find_patterns_step1})
\\
\textit{Step 2} (\S \ref{sec:find_patterns_step2}): Discover
% Find 
relevant patterns in the important heads, using criteria described in \S \ref{sec:find_patterns_step2}. 
    % (Sec.\ref{sec:find_patterns_step2})
    \\
\textit{Step 3} (\S \ref{sec:find_patterns_step3}): Evaluate and validate the patterns to confirm their global relevance. 
    % (Sec.\ref{sec:find_patterns_step3})
    %if the patterns exist in the attention heads. 

Once the important patterns are identified, there are two common approaches
% - fixed and masking -  
(i.e. fixing and masking)
to inject 
% apply
them as constraints to the attention matrices in the transformer-based neural models (see \S\ref{sec:apply_patterns}). The pipeline also 
%recommends 
enables two scenarios, 
% to apply the patterns
in which injecting
% applying
the patterns can be beneficial: 
% the first one is to enhance the original model, while the second one is to train a new model in which the patterns are enforced. 
the first one is to train a new model with the patterns injected, while the second one is to enhance the original model.
% Additional potential ways to make use of the patterns are left as future work.

\subsection{Discover Patterns from Attention}
% \subsection{Extract Patterns from Attention}
\label{subsec:extract_patterns}
In this section we provide details of the three steps for discovering
% extracting
patterns from the attention heads. 
\change{The three steps are illustrated in \autoref{fig:pipeline}} (B).

\subsubsection{Step 1: Estimate Head Importance}
\label{sec:find_patterns_step1}
Although the multi-head self attention mechanism in transformers allows the model to learn multiple types of relationships between input representations across a single hidden layer, the importance of the individual attention heads can vary depending on the downstream tasks. 
% Motivated by previous work, \citep{molchanov2019importance, michel-2019-heads}, we assume the importance of each head to be independent to avoid an NP-hard combinatorial search.
In practice, we propose the use of scalable gradient-based methods \citep{michel-2019-heads, voita-etal-2019-analyzing, molchanov2019importance} for an efficient estimation of head importance, and take the top-K heads at each layer to find important patterns for the task ~(\S\ref{sec:find_patterns_step2}). \change{Note that $K$ %is 
can be adjusted based on the availability of human users and the size of the model. 
% (i.e. set $K$ to be smaller for larger models).
} 

\subsubsection{Step 2: Find Attention
Patterns}
\label{sec:find_patterns_step2}
%In this step, the human expert should start with 
Once the the most important heads are identified, their attention distributions are inspected to look for patterns.
% the human expert visually inspects their attention distributions looking for patterns.
% Here, we define a pattern very broadly as a predicate $P$ that can be verified on any pair of input tokens $(x_i,x_j)$. 

%To start, 
We define an attention pattern to be interpretable \textit{iff} it can be modeled as a predicate $P$ between any pair of input tokens $(x_i, x_j)$.
For instance, the positional pattern `preceding token' would be true if $x_i$ appears before $x_j$. Candidate patterns can be discovered following two criteria:
% 1) be interpretable by human experts to be beneficial for the downstream tasks. 
1) they are beneficial for the downstream task;
2) they occur consistently among relevant tokens.
% \raymond{In practice, the selection criteria for any pattern is a combination of its relevance, which can be quantified by evaluating the associated predicate $P$ (Sec. \ref{sec:find_patterns_step3}) and usefulness for the task, which can be estimated by the importance score (Sec. \ref{sec:find_patterns_step1})}
% For example, in previous works on analyzing the attention heads of pretrained language models \cite{vig-belinkov-2019-analyzing, kovaleva-etal-2019-revealing}, some attention heads show a high correlation with position or linguistic properties (like syntactic dependency).
% \raymond{
% While the first criteria (beneficial) is verified by filtering out unimportant heads in Step 1, the recall for second criteria (consistency) requires the human user to explore a diverse set of examples (e.g. input length, prediction confidence) before the precision of found patterns are evaluated in Step 3.
% Lastly, it is also worth mentioning that this step of the pipeline could naturally be automated by directly evaluating the relevance and importance of predefined patterns (e.g. syntax, discourse) based on the human user's intuition for any given task. However, as 
% % evidence 
% indicated
% from the findings of our case study (Sec. \S\ref{sec:find-and-evaluate-patterns}), our human-in-the-loop approach allows users to discover interpretable patterns which otherwise would have been hard to detect automatically due to the potentially infinite search space of possible patterns.}

\subsubsection{Step 3: Evaluate Attention Patterns}
\label{sec:find_patterns_step3}
% When a specific interesting pattern is uncovered from visualizing the attention heads, the next step is to confirm its global relevance on each head by empirically measuring the proportion of total attentions from the head aligned with the pattern aggregated over data samples. By the evaluation on each single head over the whole validation set, the NLP expert can verify if the pattern generally exists across different data samples, instead of only appearing by chance on certain data that the expert happened to look at.
% From
With a pattern discovered 
% found
in \S\ref{sec:find_patterns_step2}, this step confirms the pattern's global relevance by empirically measuring the proportion of attention values aligning with the pattern. For each attention head, the associated predicate is evaluated over the entire validation set to ensure the pattern is not appearing by chance on the certain data that the user happen to look at.

Specifically, we define the global relevance (GR) of a pattern $P$ for a head $h$ as follows:
% \begin{eqnarray}
% \textrm{gr}(x,P,h) &=& \frac{\sum_{i}^{|x|}\sum_{j}^{|x|} \alpha_{i,j}^{h} \cdot \mathbbm{1}_{P(x_{i},x_{j})}}{|x|} 
% \textrm{GR}(P,h) &=&\sum_{x\in X} \frac{gr(x,P, h)}{|X|} \label{eq:gr}\\
% \end{eqnarray}
\begin{equation}
\label{eq:gr}
\textrm{GR}(P,h) = \frac{1}{|X|} \sum_{x\in X} \frac{\sum_{i}^{|x|}\sum_{j}^{|x|} \alpha_{i,j}^{(x, h)} \cdot \mathbbm{1}_{P(x_{i},x_{j})}}{|x|}
\end{equation}
where the attention value from the token $x_i$ to $x_j$ on the head $h$ for an input sample $x$, denoted as $\alpha_{i,j}^{(x,h)}$, is aggregated if and only if $P(x_{i},x_{j})$ holds. To validate a pattern's generality, the relevance is averaged over the validation set $X$.
% where $\textrm{gr}(x,P,h)$ denotes the global relevance of %\wanglj{a}
% a pattern $P$ for %\wanglj{a}
% head $h$ on a single data sample $x$, and to validate the generality, $\textrm{GR}(P,h)$ is then computed as the average $\textrm{gr}(x,P,h)$ over the validation set. The attention value from token $x_i$ to $x_j$ on the head $h$, denoted $\alpha_{i,j}^h$, is aggregated if and only if $P(x_{i},x_{j})$ holds. Note that $\sum_i^{|x|}\sum_j^{|x|} \alpha_{i,j}^{h}=|x|$ for attention matrices. 

%due to the property of attention matrices.

% Given a pattern $P$, it will be kept if there exists at least one significantly relevant head.  There are several ways to decide whether the relevant head exists based on GR, e.g., setting a threshold.  Here, we suggest to use one-tailed one-sample t-test on each head $h^*$ with the null hypothesis as: $GR(P,h^*) < \bar{GR}(P,h)$, where $\bar{GR}(P,h)$ is the average of $GR(P,h)$ over $H$. The p-value is usually set as 0.01.  If there is at least one head rejecting this null hypothesis, i.e. showing its significantly higher relevance than most other heads, we keep the pattern $P$ for further applications.

% \subsection{Apply Patterns}
\subsection{Inject Patterns}
% \footnotetext{{The figures of (a) and (b) in the `Extract Patterns' step are captured from the visual interface \cite{li-etal-2021-t3}.}} 
\label{sec:apply_patterns} 
\change{As illustrated in \autoref{fig:pipeline} (C)}, after extracting the patterns following the three steps in \S\ref{subsec:extract_patterns}, we propose to inject
% applying
the patterns into attention matrices
% transformer models 
with two methods (\S\ref{subsec:injecting_methods}), and discuss two practical scenarios (\S\ref{subsec:injecting_scenarios}) where they can be beneficial for the downstream tasks.

\subsubsection{Methods for injecting
% Applying
Patterns}
\label{subsec:injecting_methods}
% Once 
% important and interpretable 
% From the patterns extracted
% In this work, we propose applying the attention patterns
% they can be injected 
In this work, we inject
% apply 
the discovered patterns
% into a transformer model 
by either fixing or masking the attention weights prior to the softmax function. 
% Although the two methods are very similar with respect to what they can achieve, as we will see in the case study they can be more or less appropriate depending on the nature of the pattern that needs to be applied.
For fixed attention weights, the attention logits in the scaled-dot-product attention is replaced with a %dynamic
fixed (possibly input dependent) matrix %based on the input 
such that:
\begin{equation}
    \textrm{FixAttn}(V, X) = \sigma(F^{(P)}(X))V
\end{equation}
where $\sigma$ is the softmax operation, $V$ is the value vectors, and $F(X) \in [0, 1]$ computes a binary matrix from the input sequence $X$ based on the predicated $P$ for the specific pattern. %identified from our analysis. 
Similarly, a pattern can also be injected
% applied
by casting a mask over the attention weights computed from the key and query vectors, as:
\begin{equation}
    \textrm{MaskAttn}(Q,K,V,X) = \sigma(M^{(P)}(X) + QK^T)V
\end{equation}
where $M(X) \in [0, -\infty)$ computes the desired behaviour in the same fashion as $F(X)$, and is added to the attention logits to approximate the multiplication of the attention distribution by a weight. 

Although the two methods are very similar with respect to the improvement they contribute (see \S\ref{sec:case_studies}), masking allows more flexibility and is generally used for patterns with a large number of applicable tokens, while fixing is more rigid and better suited for a small number of applicable tokens.
% case study
% they can be more or less appropriate depending on the nature of the pattern that needs to be applied.

\subsubsection{Scenarios for Injecting
% Applying
Patterns}
\label{subsec:injecting_scenarios}
In practice, patterns can be injected
% applied
in at least two scenarios: (i) injecting patterns directly into the attention heads of transformer-based models, and (ii) injecting patterns into pretrained transformer models using techniques such as the Projected Attention Layers \citep{stickland2019bert}. We conduct case studies for these two scenarios in \S\ref{sec:case_studies}.

% , in what can be regarded as a human-guided distillation process. (ii) to enhance the original model on which they are discovered aiming to improve its accuracy and interpretability (
% ii) to improve smaller models, in what can be regarded as a human-guided distillation process. 
% In the first scenario, although patterns can be directly injected into the pretrained encoder, determining to which heads the patterns should be applied to requires extensive hyperparameter search and risk overfitting. Instead, we opt to inject the patterns via additional attention heads through techniques such as the Projected Attention Layers \citep{stickland2019bert}. As for the second scenario, the patterns are simply applied on the heads (one per head) for each layer, and the new models are trained from scratch.

\begin{figure*}[th!]
    \centering
    \includegraphics[width=\linewidth]{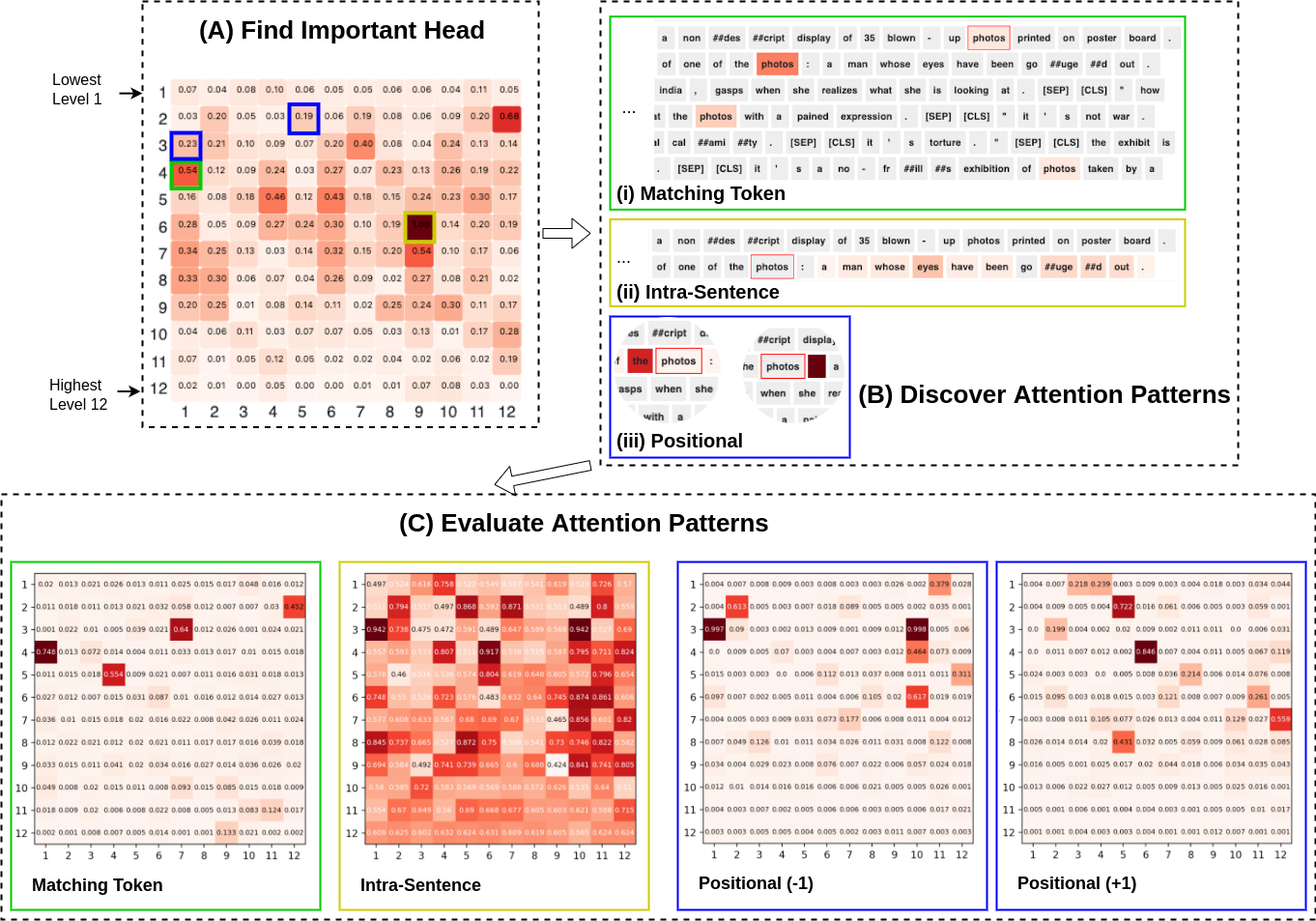}
    \caption{Example of Pattern Extraction %(middle in Fig.\ref{fig:pipeline}) 
    in the extractive summarization case study.{\footnotemark} \change{\textbf{(A)} We first find} \cut{Important} important heads \cut{are found (1)}, \change{before \textbf{(B)} identifying the three interpretable patterns (highlighted in Green, Olive and Blue, respectively): (i) Matching token, (ii) Intra-Sentence, and (iii) Positional.} \cut{Then three patterns}.  Finally, \change{\textbf{(C)}} each pattern is evaluated with the global relevance score (GR) on all of the attention heads. \cut{((c), bottom).} For \cut{better visualization} \change{the purpose of illustration}, we \cut{only label} \change{display} one attention head with significantly larger GR in \cut{(1)} for each \change{of the three identified} patterns.}
    \label{fig:example_summ}
\end{figure*}

% \section{Case Studies: Extractive Summarization}
\section{Case Studies}
\label{sec:case_studies}
In this section, 
we demonstrate the effectiveness of our pipeline in two NLP tasks (extractive summarization and topic segmentation) and discuss our findings in detail.
 
% \subsection{\cut{Trained} Model}

% \subsection{Task Models}
\subsection{Models for Tasks}
\label{subsec:case-study-models}
% % In our case study, 
% For the task of extraction summarization,
% we adopt the architecture of the popular BERTSum model \cite{liu-lapata-2019-text}
% % \footnote{We re-implemented the model using the BERT class from hugging face. Results are comparable with original paper with details in Appendix. A}.
% The model first obtains the contextualized sentence representation from the pretrained BERT encoder, and uses a single-layer binary classifier to score the sentences for summary selection. More specifically, a pair of BOS, EOS tokens are inserted before and after each sentence to indicate the segment boundary, and during prediction, the last hidden state of the BOS token of each sentence is used as the sentence representation.
% We apply our pipeline on the BERTSum model trained on the CNN/DM dataset \citep{herman-2015-teaching, see-etal-2017-get}, %which is 
% a widely used summarization dataset and as evaluation metrics we use standard ROUGE scores \citep{lin-2004-rouge}. 

We adopt the popular BERTSum \citep{liu-lapata-2019-text} for extractive summarization. With the contextualized representation from BERT, the model uses a binary classifier to predict whether each sentence belongs in the summary. We train the model on the CNN/DM dataset \citep{see-etal-2017-get}, and use ROUGE \citep{lin-2004-rouge} as the evaluation metric.

% \raymond{
% % \subsection{Topic-Segmentation Model}
% For the task of topic segmentation, we adopt the Cross-Segment BERT \cite{lukasik-etal-2020-text} for our case study. Without modification to the architecture of the Transformer encoder, Cross-Segment BERT represent each candidate sentence boundary by its left and right contexts (i.e. the sequences of word-piece tokens that come before and after, respectively, the candidate break), separated by a EOS token. The model performs independent binary classification by using the BOS token as the representation of each candidate break. For topic-segmentation experiments, we apply our pipeline on the Cross-Segment BERT model trained on the WikiSection dataset \citep{arnold-etal-2019-sector}, and 
% % the patterns discovered in the summarization model in a 
% % % inter-domain
% % cross-domain
% % setting and 
% use the Precision, Recall and F1-score as the evaluation metric. 
% }
% Align with Figure 2
\footnotetext{\textbf{(A)} and \textbf{(B)} of \autoref{fig:example_summ} are captured from the visual interface presented in \citet{li-etal-2021-t3}.}

We adopt Cross-Segment BERT \citep{lukasik-etal-2020-text} for topic segmentation, where a candidate segment boundary is first represented by its left and right context, and then passed through a binary classifier to predict whether the candidate is a topical segment boundary. The model is trained on the WikiSection dataset \citep{arnold-etal-2019-sector}, and the F1-score is used as evaluation metric for validation.

\subsection{Discover Patterns from Attentions}
% \subsection{Extract Patterns from Attentions}
% We begin the pipeline by performing the extraction process from the two task models described in \S\ref{subsec:case-study-models}. 
Using the two models from \S\ref{subsec:case-study-models}, as we discover
% found
that similar attention patterns exist in the important heads for both tasks, the two case studies are presented together. 
% An overview of the process for extractive summarization is illustrated in \autoref{fig:example_summ}.
%We will reference the overview of the extractive summarization process~\autoref{fig:example_summ} as a running example throughout this section. 
\cut{Due to the lack of space} \change{Without loss of generality}, \cut{next} we will use extractive summarization as the running example task (\autoref{fig:example_summ}) to illustrate the process of pattern discovery. \cut{similar} We also apply the same process \cut{is applied} to topic segmentation. \cut{as well}
%In summary, a
% As we found that similar attention patterns exist in the important heads for both tasks, the two case studies are presented and discussed together. 
% As an overview, the process of extracting patterns for the extractive summarization case study is shown in \autoref{fig:example_summ}. 

%\vspace{-0.3ex}
\subsubsection{Find Important Heads}
\label{sec:find-important-heads}
% To measure the importance of the attention head for the task of extractive summarization
We adapt the Taylor expansion method \citep{molchanov2019importance} as a proxy score for the head importance estimation. 
% in our case studies.
% w.r.t. the loss function.
% Specifically, the method was proposed 
% % by \citet{molchanov2019importance} 
% to estimate the error induced from removing a parameter from the model. 
Following \citet{li-etal-2021-t3}, we use the first-order expansion to avoid the overhead from computing the Hessian, where the gradient w.r.t. the validation loss is summed over all parameters of an attention head to estimate its importance.

% \cut{
% To identify the most appropriate head importance estimation method for extractive summarization, we evaluate the three proxy scores (Sensitivity, LRP and Taylor) against the leave-one-out head importance score by using the selection process outlined in \S\ref{sec:find_patterns_step1}. As shown in Table \ref{tab:estimation-similarity}, the Taylor Estimation \citep{molchanov2019importance} is the most aligned with leave-one-out in term of cosine similarity and therefore it is used in the the rest of our analysis.
% }

% \begin{table}[h!]
%     \centering
%     \resizebox{0.9\linewidth}{!}{
%     \begin{tabular}{l|ccc}
%     Method  & Sensitivity & LRP & Taylor \\
%     \hline
%     Cosine Similarity  & $0.6$ & $0.64$ & $\mathbf{0.83}$\\
%     \end{tabular}}
%     \caption{Cosine similarity between the leave-one-out loss increment and the three estimation methods.}
%     \label{tab:estimation-similarity}
% \end{table}

The 
% resulting estimated 
importance score heatmap of all heads is visualized in \autoref{fig:example_summ} (A), revealing that head importance is not uniformly distributed, i.e. a small number of heads play a dominant role for the summarization task, 
%which is in line with the findings by \citet{michel-2019-heads}.
as observed in \citet{michel-2019-heads}.

% \subsubsection{Finding and Evaluating Patterns}
\subsubsection{Discover and Evaluate Patterns}
\label{sec:find-and-evaluate-patterns}

To discover 
% find
task-specific patterns,
we analyze the top-3 most important heads of each layer, and look for
% looking for
human-interpretable relationships encoded in the attention weights. In practice, we use the instance-level interactions provided by the visual framework \citep{li-etal-2021-t3}, and randomly select 5 validation examples per task for our analysis.
% select a handful of examples (3-5) from the validation for our analysis.
The entire process takes
% took
less than one hour to complete for each task, where we manually examine
% examined
the attention weights for less than half of the tokens for each example.
% ($\approx 200$). 
It is worth noting   that detailed analysis regarding the trade-off between human cost 
and
% vs.
pattern recall 
% as well as the consistency
% /disagreement
% among human 
% NLP experts
would require extensive user studies beyond the scope of this work. 

% From the pattern finding process, three specific types of patterns appear to be present in the most important heads for both tasks. \raymond{This is consistent with the findings from \citet{bian-etal-2021-attention}, where the roles of attention heads do not differ significantly across tasks, and suggest the patterns are generalizable across multiple tasks. We hope our findings provide motivation for future work in studying the differences between the importance of attention patterns across models and tasks}. 
% Those patterns are then evaluated on the validation set to assess their global relevance on each head.

% From the discovered patterns,
% in the top-3 most important heads of each layer
% we then evaluate them on the validation set to assess their global relevance on each head. 
After discovering patterns,
% the pattern discovering
% finding
% process
we assess the global relevance of each patterns on the validation set, where
% Specifically, we keep 
the pattern is kept only if the corresponding predicate $P$ exists in at least one significantly relevant head. 
% There are several ways to decide whether the relevant head exists based on GR, e.g., setting a threshold.
In our case studies, we use the 3-sigma rule
% test 
% to validate patterns
to determine the significance of a pattern. Specifically, patterns with at least one head over 3 standard deviations above the GR mean (over all the heads) are kept for further applications. 
% we use the one-tailed one-sample t-test on each head $h^*$ with the null hypothesis as: $GR(P,h^*) < \bar{GR}(P,h)$, where $\bar{GR}(P,h)$ is the average of $GR(P,h)$ over $H$. The p-value is usually set as 0.01 \citep{fisher1992statistical}.  If there is at least one head rejecting this null hypothesis, i.e. showing its significantly higher relevance than most other heads, we keep the pattern $P$ for further applications.

% From the pattern finding process, three specific types of patterns appear to be present in the most important heads for both tasks. This is consistent with the findings from \citet{bian-etal-2021-attention}, where the roles of attention heads do not differ significantly across tasks, and suggest the patterns are generalizable across multiple tasks. We hope our findings provide motivation for future work in studying the differences between the importance of attention patterns across models and tasks

After verifying on the validation set, we discover 
three patterns 
% that
consistently existing in both tasks (over 50\% of important heads). 
This suggests that important patterns are generalizable across multiple NLP tasks, which
% and
is consistent with the findings in \citet{bian-etal-2021-attention}. \change{Further analysis also shows that the attention patterns are consistent after fine-tuning, where we report an average Jensen-Shannon Divergence 
% divergence
of $0.01$ between the attention distributions of BERTSum across 3 random seeds.}
% This is consistent with the findings from \citet{bian-etal-2021-attention}, 
% % where the roles of attention heads do not differ significantly across tasks,
% and suggests that important patterns are generalizable across multiple tasks.
We hope our findings provide motivation for the in-depth study of pattern importance in different NLP tasks.  Lastly, while it may be argued that this step of the pipeline can be automated by directly evaluating the importance and relevance of predefined patterns (e.g. syntax, discourse) based on intuitions, as indicated below, our interactive approach allows the discovery of interpretable patterns which otherwise would be hard to define due to the infinite search space of possible patterns. Next, we describe the three discovered patterns in detail.

% Lastly, 
% % we mentioning that 
% it could be argued that this step of the pipeline could naturally be automated by directly evaluating the relevance and importance of predefined patterns (e.g. syntax, discourse) based on the human user's intuition for any given task. However, as 
% % evidence 
% indicated from the findings described below, 
% % (Sec. \S\ref{sec:find-and-evaluate-patterns}),
% our human-in-the-loop approach allows users to discover interpretable patterns which otherwise would have been hard to detect automatically due to the potentially infinite search space of possible patterns.

\paragraph{Matching Token (Green in \autoref{fig:example_summ})}
% We observe that the attention weights of some important heads appear to exhibit an ``attending to matching tokens'' pattern.
This pattern describes the ``attending to matching tokens'' behaviour, where
% Specifically, 
the attention value $\alpha_{i,j}^{h}$ between input tokens $x_i$ and $x_j$ on the head $h$ is high whenever $x_i = x_j$. For example, as shown in \autoref{fig:example_summ} (i), the token "photo" mostly attends to other appearances of the token "photo" in the input sequence. To evaluate whether this pattern has a large global relevance for any head, we only consider tokens that appear at least twice within a single documents, and compute GR (Eq. \ref{eq:gr}), in which $P(x_i,x_j)$ holds if and only if $x_i=x_j$, i.e. $\mathbbm{1}_{P(x_i,x_j)} = (\mathbbm{1}_{\textrm{freq}(x_i) > 1}) \times (\mathbbm{1}_{x_i=x_j})$. 

The evaluation results show that there are several heads for which the matching token pattern has high global relevance 
% with 
(See the Green box in \autoref{fig:example_summ}). 
% Interestingly, these heads are prominent in the importance heatmap, which suggests the matching token pattern is critical for the summarization task. 
Interestingly, these heads are more prominent (in the importance heatmap) for the extractive summarization task, suggesting this pattern is especially important for summarization models during inference.

\paragraph{Intra-Sentence/Context (Olive in  \autoref{fig:example_summ})}
% For some heads, the attentions among tokens tend to be localized within the sentence boundaries, as shown on an example in Figure.\ref{fig:example_summ} (b). 

% For some heads, the attentions among tokens tend to be localized within pre-defined boundaries. 
This pattern describes the behaviour of only attending to tokens within a text span.
For summarization, these heads will focus on attending tokens within the same sentence (\autoref{fig:example_summ} (ii)). \cut{And} Similarly, the same heads in topic segmentation models will focus on attending tokens within the same context (left or right).
To evaluate this pattern, GR is computed with $P(x_i,x_j)$ holding \textit{iff} $x_i$ and $x_j$ occur within the same text span. \autoref{fig:example_summ}~(C) reveals that this pattern appears more frequently in the mid to upper layers of the transformer encoder.
% \raymond{We 
% % found
% find these heads to be more important for topic segmentation models.}

\paragraph{Positional (Blue in \autoref{fig:example_summ})}
Similar to 
\citet{kovaleva-etal-2019-revealing}, we also observe ``positional heads'', which focus specifically on either the preceding or following tokens, i.e., either $\alpha_{i,i-1}^{h}$ or $\alpha_{i,i+1}^{h}$ have high values (\autoref{fig:example_summ} (iii)). To evaluate this pattern, GR is computed with $P(x_i,x_j)$ holding \textit{iff} $j=i-1$ for preceding postional heads and $j = i + 1$ for succeeding positional heads. The pattern is verified to exist in the lower layers of the encoder, shown in the blue boxes of \autoref{fig:example_summ} (C). 

\change{\paragraph{Other Patterns} In addition to the three patterns mentioned above, we also observe heads that focus on attending to special tokens (e.g. [CLS], [SEP]) or punctuations (e.g. periods). However, we find that attention heads with this behaviour are generally less important for the task (outside top-3), and therefore omitted them from the next step of our pipeline. 

On the other hand, we also find uninterpretable attention patterns in some of the important heads of each layer. As hypothesized by previous works \citep{clark-etal-2019-bert}, these attention heads might be performing complex linguistic operations in combination with other heads. We leave the verification, interpretation and the efficient injection of these patterns into the models as a direction for future work.}

% \raymond{\subsection{Discussion on Human Factors}
% As with all human-in-the-loop approaches, our pipeline is dependent on NLP experts to identify important patterns for the task. While the associated human cost and consistency/disagreement among NLP experts are important factors to consider, this would require extensive user studies beyond the scope of this paper. Nevertheless, something that we will highlight is that the pattern extraction process can be naturally automated by evaluating for complex patterns (e.g. discourse, syntax, semantics) based on their ``gold-annotation'' and injecting them based on their estimated importance (as a hyperparameter) to the final model. More tellingly, our case study interestingly shows that, with a human-in-the-loop approach, human experts can discover useful patterns (i.e. matching tokens) which might be hard to detect automatically due to the large search space of potential patterns. Finally, with regard to the human cost, our case study on extractive summarization task took two NLP experts less than one hour with help from the visualization framework.}

\begin{table*}[ht]
\centering
\resizebox{\linewidth}{!}{
\begin{tabular}{lccccccccccccc}
\hline
\multicolumn{1}{c|}{\multirow{3}{*}{\large Model}} &
  \multicolumn{1}{c|}{\multirow{3}{*}{\large Sparsity ($\rho$)}} &
  \multicolumn{6}{c|}{\large Extractive Summarization} &
  \multicolumn{6}{c}{\large Topic Segmentation} \\ \cline{3-14} 
\multicolumn{1}{c|}{} &
  \multicolumn{1}{c|}{} &
  \multicolumn{3}{c|}{CNN/DM} &
  \multicolumn{3}{c|}{NYT-50} &
  \multicolumn{3}{c|}{WikiSection} &
  \multicolumn{3}{c}{Wiki-727K} \\ \cline{3-14} 
\multicolumn{1}{c|}{} &
  \multicolumn{1}{c|}{} &
  R-1 &
  R-2 &
  \multicolumn{1}{c|}{R-L} &
  R-1 &
  R-2 &
  \multicolumn{1}{c|}{R-L} &
  P &
  R &
  \multicolumn{1}{c|}{F-1} &
  P &
  R &
  F-1 \\ \hline
\rowcolor{Gray} \multicolumn{14}{c}{6 Layer 8 Heads} \\ \hline
\multicolumn{1}{l|}{Transformer} &
  \multicolumn{1}{c|}{0} &
  40.50 &
  18.22 &
  \multicolumn{1}{c|}{36.94} &
  45.86 &
  26.83 &
  \multicolumn{1}{c|}{38.23} &
  0.698 &
  0.647 &
  \multicolumn{1}{c|}{0.671} &
  0.647 &
  0.243 &
  0.353 \\ \hline
\multicolumn{1}{l|}{+Patterns (4/8)} &
  \multicolumn{1}{c|}{0.43} &
  \textbf{41.42} &
  \textbf{18.94} &
  \multicolumn{1}{c|}{\textbf{37.92}} &
  47.11 &
  27.89 &
  \multicolumn{1}{c|}{39.34} &
  0.744 &
  \textbf{0.711} &
  \multicolumn{1}{c|}{\textbf{0.727}} &
  0.624 &
  \textbf{0.318} &
  \textbf{0.421} \\ \hline
\multicolumn{1}{l|}{+Patterns (8/8)} &
  \multicolumn{1}{c|}{0.86} &
  41.37 &
  18.92 &
  \multicolumn{1}{c|}{37.89} &
  \textbf{47.17} &
  \textbf{27.94} &
  \multicolumn{1}{c|}{\textbf{39.37}} &
  \textbf{0.771} &
  0.666 &
  \multicolumn{1}{c|}{0.714} &
  \textbf{0.668} &
  0.274 &
  0.395 \\ \hline
\rowcolor{Gray} \multicolumn{14}{c}{6 Layer 12 Heads} \\ \hline
\multicolumn{1}{l|}{Transformer} &
  \multicolumn{1}{c|}{0} &
  40.53 &
  18.22 &
  \multicolumn{1}{c|}{36.98} &
  46.07 &
  27.01 &
  \multicolumn{1}{c|}{38.42} &
  0.681 &
  0.680 &
  \multicolumn{1}{c|}{0.681} &
  0.687 &
  0.255 &
  0.372 \\ \hline
\multicolumn{1}{l|}{+Patterns (4/12)} &
  \multicolumn{1}{c|}{0.29} &
  41.58 &
  19.10 &
  \multicolumn{1}{c|}{38.10} &
  46.84 &
  27.68 &
  \multicolumn{1}{c|}{39.08} &
  0.752 &
  \textbf{0.717} &
  \multicolumn{1}{c|}{\textbf{0.734}} &
  0.643 &
  \textbf{0.350} &
  \textbf{0.453} \\ \hline
\multicolumn{1}{l|}{+Patterns (8/12)} &
  \multicolumn{1}{c|}{0.58} &
  41.66 &
  19.17 &
  \multicolumn{1}{c|}{38.17} &
  47.15 &
  \textbf{27.95} &
  \multicolumn{1}{c|}{\textbf{39.38}} &
  \textbf{0.757} &
  0.701 &
  \multicolumn{1}{c|}{0.730} &
  0.655 &
  0.342 &
  0.450 \\ \hline
\multicolumn{1}{l|}{+Patterns (12/12)} &
  \multicolumn{1}{c|}{0.86} &
  \textbf{41.68} &
  \textbf{19.16} &
  \multicolumn{1}{c|}{\textbf{38.19}} &
  \textbf{47.17} &
  27.94 &
  \multicolumn{1}{c|}{\textbf{39.38}} &
  0.756 &
  0.702 &
  \multicolumn{1}{c|}{0.728} &
  \textbf{0.663} &
  0.318 &
  0.430 \\ \hline
\rowcolor{Gray} \multicolumn{14}{c}{6 Layer 12 Heads (with BERT Embeddings)} \\ \hline
\multicolumn{1}{l|}{Transformer} &
  \multicolumn{1}{c|}{0} &
  40.74 &
  18.40 &
  \multicolumn{1}{c|}{37.20} &
  46.07 &
  26.98 &
  \multicolumn{1}{c|}{38.37} &
  0.738 &
  0.674 &
  \multicolumn{1}{c|}{0.704} &
  0.665 &
  0.415 &
  0.511 \\ \hline
\multicolumn{1}{l|}{+Patterns (4/12)} &
  \multicolumn{1}{c|}{0.29} &
  41.49 &
  19.07 &
  \multicolumn{1}{c|}{37.99} &
  47.02 &
  27.83 &
  \multicolumn{1}{c|}{39.21} &
  \textbf{0.782} &
  0.715 &
  \multicolumn{1}{c|}{0.747} &
  \textbf{0.674} &
  \textbf{0.423} &
  \textbf{0.520} \\ \hline
\multicolumn{1}{l|}{+Patterns (8/12)} &
  \multicolumn{1}{c|}{0.58} &
  41.57 &
  19.11 &
  \multicolumn{1}{c|}{38.08} &
  47.16 &
  \textbf{27.96} &
  \multicolumn{1}{c|}{\textbf{39.40}} &
  0.760 &
  \textbf{0.737} &
  \multicolumn{1}{c|}{\textbf{0.748}} &
  0.665 &
  0.421 &
  0.515 \\ \hline
\multicolumn{1}{l|}{+Patterns (12/12)} &
  \multicolumn{1}{c|}{0.86} &
  \textbf{41.61} &
  \textbf{19.15} &
  \multicolumn{1}{c|}{\textbf{38.14}} &
  \textbf{47.17} &
  27.95 &
  \multicolumn{1}{c|}{39.37} &
  0.761 &
  0.731 &
  \multicolumn{1}{c|}{0.745} &
  0.666 &
  0.367 &
  0.473 \\ \hline
\end{tabular}
}
\caption{Results for the two tasks (four datasets) under different settings, \change{where we report the average performance across the top-3 checkpoints}. The parenthesis (e.g. 4/8) denotes the number of heads with patterns injected, while sparsity ($\rho$) is computed from the average of the 4 datasets. 
%     % - as in the original transformer \cite{transformer} (6 layers, 8 heads), and as in the distilled models \cite{Sanh2019DistilBERTAD, jiao-etal-2020-tinybert}(6 layers, 12 heads w/ or w/o BERT Embeddings)
}
\label{tab:sparse-transformer}
\end{table*}

% \subsection{Applying Patterns to Models}
\subsection{Injecting Patterns to Models}

After uncovering potentially important patterns %encoded in the attention heads 
and confirming their relevance, %existence through evaluating on the validation, we justify their benefits for the task of extractive summarization by applying 
we inject 
% apply
them to transformer-based models 
% \cut{summarizers}
for the 
% task of summarization (intra-domain) and topic-segmentation (inter-domain)} 
task of summarization 
% (both in-dataset and cross-dataset)
and topic segmentation
through masking and fixing the attention weights.
While we only perform the pattern discovery
% \textit{extraction}
process on the CNN/DM and WikiSection datasets, we inject
% apply
the discovered patterns to two other datasets (NYT-50 \citep{sandhaus2008new} for summarization and Wiki-727K \citep{arnold-etal-2019-sector} for topic segmentation) to demonstrate that our discovered patterns are generalizable in ``cross-dataset'' settings\footnote{Results shown in Sec. 4 are without the Trigram Blocking trick, and more results \textit{with} it are in Appendix~\ref{sec:trigram}
% \cut{
% Aiming for more general `cross-dataset' and `cross-task` insights, 
% %Notice that 
% in addition to the CNN/DM dataset, %which is the dataset to 
% on which we trained the model that we use to extract the patterns, we also evaluate the %functionality 
% benefit of the patterns when applied to the
% summarization dataset of
% NYT-50
% % dataset
% \citep{sandhaus2008new} 
% as well as two topic segmentation datasets, namely, Wiki-727K \citep{koshorek-etal-2018-text} and WikiSection \citep{arnold-etal-2019-sector}.
% %, xu-durrett-2019-neural}. %, which can be regarded as a `cross-dataset' experiment.
% }
% \ref{sec:trigram}
}.

\subsubsection{
% Method
Method for Fixing and Masking
}
%In order to apply 
The patterns identified from our analysis can be injected 
% applied 
into an attention head through masking or fixing its corresponding attention weight matrix. Specifically, for the matching token pattern, we apply an attention mask which enforces that when a token  appears more than once in the document, it should attend only to other occurrences of itself:
\begin{equation}\label{eq:match_mask}
M^{(m)}_{i,j} = 
\begin{cases} 
1 & (x_i = x_j) \lor (\textrm{freq}(x_i) = 1)\\
0 & \textrm{otherwise}
\end{cases}
\end{equation}
where the constraint is removed for tokens occurring only once in the document.

Similarly, for intra-sentence/intra-context attention, the attention mask specifies that only tokens within the same boundary can attend to each others, where:
% \vspace{-1em}
\begin{equation}\label{eq:intra_mask}
M^{(s)}_{i,j} = 
\begin{cases} 
% 1 & \textrm{SameSent}(x_i, x_j) \\
1 & \textrm{SameBoundary}(x_i, x_j) \\
0 & \textrm{otherwise}
\end{cases}
\end{equation}

Lastly, we use a fixed attention matrix to encode the two positional patterns with:
\begin{equation}\label{eq:head_mask}
F^{(-1)}_{i,j} = 
\begin{cases} 
1 & j = i - 1 \\
0 & \textrm{otherwise}
\end{cases}
\end{equation}
% \vspace{-.7em}

% And 
With $F^{(+1)}_{i,j}$ being the same, but equal to 1 for $j = i + 1 $. We use fixed attention matrices %$F$ 
for these patterns to save computational overhead since it has the same effect as applying the mask (each row is a one-hot vector). This is similar to the method proposed by \citet{raganato-etal-2020-fixed}, but we only fix for the preceding and succeeding
% following
token patterns.

\subsubsection{Pattern-Infused Sparse Transformers}
\label{sec:kd}

In the first round of experiments, we inject
% apply
the four patterns
% three types of patterns
on smaller transformer models to demonstrate their effectiveness
% usefulness
on both tasks. Since the goal of these experiments is to assess the benefits brought by these patterns, we do not perform extensive hyper-parameter search when injecting
% applying
these patterns (e.g. on which layer, etc.). 
% The 

Under both settings, each of the four patterns (including two positional patterns) is injected
% applied
in a separate attention head across all layers in the model. \raymond{Motivated by \cut{previous} studies on the trade-off between sparsity ratio and task performance, we adopt the sparsity ratio used by previous works \citep{shi2021sparsebert, wang-etal-2022-learning}: $\rho = 1 - |M|/N^2$, where $|M|$ denotes the number of non-zero elements in the attention mask, and $N$ denotes the length of the example. Given the sparsity $\rho$, the complexity of self-attention is thus reduced to $\mathcal{O}\big((1-\rho)n^2\big)$ \citep{shi2021sparsebert}.
To investigate how the sparsity ratio affects the performance of our pattern-infused models, we experiment with different number of heads to inject
% apply
our patterns, where the sparsity ratio increases along with the number of heads (with patterns).}
% Note that since the goal of these experiments is to assess the benefits of the patterns, we do not perform extensive hyperparameters search when applying these patterns (e.g. on which layer, on how many heads, etc.). 

% \cut{
% As shown in Table \ref{tab:kd-summ}, for CNN/DM the pattern-infused models significantly outperform the baseline vanilla summarizer under all three settings (6-8, 6-12, and 6-12 w/ BERT embeddings), and can also beat the two distilled models with the same setting, which convincingly demonstrates the utility of applied patterns. As for the cross-dataset experiments on the NYT-50 dataset, our models significantly improves over the vanilla transformer and DistilBERT, but lags behind TinyBERT in performace. Since TinyBERT also distilled the layer output along with the attention weights and predicted distribution of BERT, it has a substantial advantage over our technique that only distils attention patterns. We suspect that the additional knowledge helped TinyBERT in capturing long term dependencies required by NYT-50.
% }

As shown in \autoref{tab:sparse-transformer}, our pattern-infused models outperform the plain transformer models for both the CNN/DM and NYT-50 datasets under all three settings
% (6-8, 6-12, and 6-12 w/ BERT embeddings)
(6 Layer 8 Heads, 6 Layer 12 Heads, and 6 Layer 12 Heads with BERT embeddings).
Similarly for topic segmentation, 
%with the exception of recall score under the 6-8 model setting, 
results also show that the pattern-injection approach substantially outperforms the vanilla transformer across all metrics.
% \change{Note that ROUGE is not on the same scale of F-1, where the ROUGE scores of oracle (best possible) extractive summaries are 52.59/31.24/48.87 for CNNDM and 49.18/33.24/46.02 for NYT.}
\change{It is worth 
% mentioning 
emphasizing that the performance gain is slightly higher for summarization models. When normalized by the ROUGE scores of extractive oracle summaries\footnote{As reported by \citet{liu-lapata-2019-text}, the ROUGE scores (R-1/R-2/R-L) of the oracle upper bound for CNN/DM and NYT-50 are respectively, 52.59/31.24/48.87 and 49.18/33.24/46.02.}, the pattern-infused summarization models achieve an average %of 
$15\%$ improvement over the baselines, while the topic-segmentation models achieve a $12\%$ improvement over the baselines.}
\change{In-line with prior work \citep{mccoy-etal-2020-berts}, we also find that the performance is consistent across random seeds, where we report an extremely low standard deviation of 
% only
0.03 (ROUGE) and 0.002 (F1) for extractive summarization and topic segmentation, respectively.  Overall, the} \cut{The} results from our experiments convincingly demonstrates the benefits of our approach and the generalizability of the patterns discovered by our pipeline. 

\raymond{In addition, while a higher sparsity ratio causes a slight decrease in performance under some scenarios, we find that even with a ratio of $0.86$,
% we were surprised to learn that even with a ratio of $0.86$,
our model still significantly outperforms the vanilla transformer across all settings. This is in contrast to the findings by previous work \citep{child2019generating, guo-etal-2019-star, li2019enhancing, Beltagy2020Longformer, zaheer2020big, shi2021sparsebert}, where the high sparsity ratio from fixed patterns often results in performance degradation from the vanilla transformer. These findings from our work provide crucial insights for designing more energy efficient models in the future.
}

% Overall, with the specific patterns applied, our models are arguably more interpretable than both 
% % vanilla
% plain transformers, as we know with certainty
% % certainly know 
% the information encoded in each masked/fixed attention heads. To further justify our claim of interpretability, the attention heads expressing the patterns tend to have higher importance scores than the other heads \footnote{An illustrative example is shown in Appendix. B.2}, suggesting that such patterns are effectively leveraged by the model. 

\begin{table}[t]
    \centering
    \resizebox{0.7\linewidth}{!}{
    \begin{tabular}{l|c|c|c}
     Model & R-1 & R-2 & R-L\\
    \hline
    Transformer &40.50&18.22&36.94\\
    \quad + match (m)&+0.03&+0.12&+0.07\\
    \quad + intra (i)&+0.05&+0.06&+0.12\\
    \quad + pos (p) &-0.16&-0.17&-0.13\\
    \quad + m +i& +0.84&+0.65&+0.91\\
    \quad + m +p&+0.07&+0.11&+0.11\\
    \quad + i + p&-0.01&+0.03&+0.07\\
    \quad + all& \textbf{+0.92}&\textbf{+0.72}&\textbf{+0.98}\\
    \end{tabular}}
    \caption{Ablation study on the CNN/DM dataset
    with the \cut{basic} \change{6 Layer 8 Head} transformer setting.
    }
    \label{tab:kd-ablation}
\vspace{-1em}
\end{table}

Overall, with the discovered patterns injected,
% applied
our models are arguably more interpretable than 
% vanilla
plain transformers on both tasks, as we know with certainty
% certainly know 
the information encoded in each masked/fixed attention heads. To further justify our claim of interpretability, the attention heads 
with patterns injected
tend to have higher importance scores than the other heads\footnote{An illustrative example is shown in Appendix~\ref{sec:distilled-importance}}, suggesting that such patterns are effectively leveraged by the model.

To study the contribution of  individual patterns, we perform an ablation study by injecting
% applying
all combinations of patterns on CNN/DM using the transformer model with 6 layers and 8 heads\footnote{Ablation study results for topic segmentation (WikiSection) can be found in Appendix \ref{sec:topic-ablation}}. 
% According to \autoref{tab:kd-ablation}, 
From \autoref{tab:kd-ablation},
we observe that injecting
% applying
matching token and intra-sentence together achieves the strongest improvement in accuracy 
% on the performance 
among all combinations, only slightly lower than injecting
% applying
all patterns. Meanwhile, the gains from injecting
% applying
patterns separately are only marginal. One intriguing explanation is that these two patterns allows the model to learn sentence-level features based on term frequency (plausibly similar to TF-IDF \citep{jones1972statistical}), where higher scores are assigned to sentences containing frequently appearing tokens. Additionally, although injecting
% applying
\textit{only} the positional patterns causes the performance to degrade, it works better when combined with the two other patterns. 
% The reason why this happens is unclear and further study are left as future work.
We 
% hypothesis 
hypothesize that 
% (local) 
positional patterns need to be combined with patterns with more global context in order 
% for the patterns
to be more effectively utilized.

\begin{table}[!ht]
    \centering
    \resizebox{\linewidth}{!}{
    \begin{tabular}{l|c|c|c|c|c|c}
    \multirow{2}{*}{Model} & \multicolumn{3}{c|}{CNN/DM (in-dataset)} & \multicolumn{3}{c}{NYT-50 (cross-dataset)} \\
    \cline{2-7}
    & R-1 & R-2 & R-L & R-1 & R-2 & R-L \\
    \hline
    BERTSum & 42.33 & 19.88 & 38.86 & 48.37 & 29.25 & 40.72 \\
    + PAL & 42.34 & 19.88 & 38.86 & 48.56 & 29.41 & 40.91 \\
    + PAL + Patterns & \textbf{42.58} & \textbf{20.05} &  \textbf{39.10} & \textbf{48.74} & \textbf{29.60} & \textbf{41.11}\\
    \end{tabular}}
    \caption{ROUGE F-scores of PAL with pretrained models for extractive summarization. \change{All metrics were significantly better than the baselines with a confidence level of $99\%$ according to the Bootstrap Significance test \citep{dror-etal-2018-hitchhikers}.}}
    \label{tab:pal-summ}
    % \vspace{-.7em}
\end{table}

% \begin{table}[]
%     \centering
%     \resizebox{\linewidth}{!}{
%     \begin{tabular}{l|c|c|c|c|c|c}
%     \multirow{2}{*}{Model} & \multicolumn{3}{c|}{WikiSection} & \multicolumn{3}{c}{Wiki-727K} \\
%     & P & R & F-1 & P & R & F-1 \\
%     \hline
%     Cross-Seg BERT & 0.830 & \textbf{0.856} & 0.805 & 0.645 & \textbf{0.746} & 0.568 \\
%     + PAL & 0.829 & 0.847 & \textbf{0.812} & 0.645 & 0.743 & 0.571 \\
%     + PAL + Patterns & \textbf{0.832} & 0.854 & 0.809 & \textbf{0.646} & 0.739 & \textbf{0.575}\\
%     \end{tabular}}
%     \caption{Precision, Recall, and F1-scores of PAL with pretrained models for topic segmentation.}
%     \label{tab:pal-ts}
% \end{table}
 
\subsubsection{Guided Pattern Injection into Pre-trained Models}

We then experiment with injecting the patterns back into the pre-trained \raymond{transformer encoder}.
% \cut{BERTSum summarizer}
In particular, we inject
% apply
them through additional attention heads in the form of a Projected Attention Layer (PAL) \citep{stickland2019bert}, along with the parameters of the original model. \raymond{Details regarding PALs are described in Appendix \ref{sec:pal-details}}.

The hidden size of our PALs is $256$, which consists of $4$ additional attention heads ($d_k=d_v=d_q=64$). PAL is added in each of the 12 BERT layers, where our patterns are injected
% applied
in the 4 PAL attention heads. To ensure the changes in performance are due to the patterns rather than the additional parameters, we also compare against adding  PAL without injecting
% applying
the patterns.

Results in \autoref{tab:pal-summ} indicate that injecting
% applying
the patterns in PAL (+PAL+Patterns) surprisingly improves BERTSum's performance on both datasets, where the performance gains on the NYT-50 are similar (or even slightly better) than on the in-domain CNN/DM dataset, supporting the generality of the discovered patterns. 
% This suggests that following our pipeline can boost model performance, as well as its interpretability, as the model follows meaningful patterns. 
% \raymond{However, we do not see a significant improvement for the topic segmentation models, we hypothesize that since the patterns are extracted from summarization models, their marginal benefits are less significant for BERT, and task-specific patterns might be needed (I don't have a good justification, consider removing this results).}
Additionally, as it was the case for the 
% pattern-infused
transformers
with patterns injected, visualizing the head importance scores reveals that the PAL heads with patterns injected 
% applied
are significantly more important (by two orders of magnitude) than the PAL heads without patterns injected\footnote{An illustrative example is shown in Appendix~\ref{sec:pal-importance}
% applied
% \ref{sec:pal-importance}
}, \raymond{indicating that the interpretable patterns are important features during model inference}.

\raymond{
In summary, the key aim of our experiments was to verify consistent improvements over our own baselines under the same settings in order to probe 
%in a controlled environment 
the benefits (effectiveness and efficiency) of the discovered patterns for the task. Therefore, we do not perform extensive tuning to achieve the same results reported by \citet{liu-lapata-2019-text}.
% It is worth mentioning that although we did not achieve higher ROUGE score than the ones reported by \citet{liu-lapata-2019-text}, the key aim of our experiments was to verify consistent improvements over our own baselines under the same settings, in order to probe in a controlled environment the benefits (task performance and efficiency) of the discovered patterns for the task.
}

\section{Conclusion and Future Work}

In this paper, we propose a generic human-in-the-loop pipeline, which combines 
% synergizes
two popular research directions, where the findings from an  analysis of the multi-head self-attention mechanism in transformers can be utilized to create more accurate and interpretable  transformer models. %To be specific, a
A human 
% expert
analyzes the attention heads of a task-specific model, discovers and verifies potentially meaningful patterns, and injects them into the attention heads of models. 
% By running a case study on the extractive summarization and topic segmentation tasks
By running a case study on two NLP tasks, we show the effectiveness of our pipeline. We do discover
% find
meaningful patterns in some important heads, and the relationships encoded in the patterns help us understand the features used by the model for both tasks.
% when performing summarization and topic segmentation.
Furthermore, by injecting
% applying
the patterns into the smaller models and the original model, the performance and interpretability get improved in both cases.  

As future work, we plan to apply our pipeline to other NLP tasks (e.g. language modeling, abstractive summarization) and explore and verify whether the important patterns from one task can be transferable to another task. \change{Similarly, we also plan to apply our pipeline to different model variants to examine and compare the patterns encoded in the attention weights.}  
% More 
In the long term, \change{
% our pipeline could be naturally automated by replacing the pattern discovery step with predefined linguistic patterns, however the efficiency gains from injecting such patterns (requiring ground-truth annotations) would require more in-depth studies beyond the scope of this paper.
our pipeline could be naturally automated by replacing the pattern discovery step with evaluating predefined linguistic patterns. However, assessing the efficiency gains from injecting such patterns (requiring ground-truth annotations) would require more in-depth studies beyond the scope of this paper.
Finally,} since human factors are an important aspect of interpretability, we plan to conduct extensive user studies \change{across different NLP tasks and model sizes} to examine the trade-off between human-cost and the coverage of discovered patterns. 

% One very promising direction for future work is to apply our generic pipeline to other NLP tasks, like machine translation, whose models are known to contain redundancy among the heads \cite{michel-2019-heads, voita-etal-2019-analyzing}. It would also be worth exploring and verifying whether the important patterns from one task can be transferable to another task (e.g. extractive summarization to abstractive summarization, or to discourse parsing, etc.), in order to better understand the connections between different tasks. In addition to the apparent patterns that human experts can identify through visualization of the attention, our pipeline can be further expanded to consider deeper linguistic features, i.e. human experts could propose linguistic patterns  that may be relevant for a task based on prior knowledge (e.g. discourse may be important for summarization), and then evaluate those patterns, to apply them back to the model once validated. 

% Does not count towards page limit
\section*{Limitations}
The scope of our case studies is limited to English datasets consisting of long documents for BERT-based models. Additionally, we only adopt the visual interface proposed by \citet{li-etal-2021-t3} due to its support for long documents, and leave the design and implementation of additional visualization techniques as a venue for future work. 
% Further, we do not perform extensive user studies to examine human cost and agreement between users involved in the pattern discovery process. Lastly, due to the limitation of computational resources, we do not perform large-scale pretraining with the discovered patterns, and only present evidence for its feasibility from the results in our case studies.  

\section*{Acknowledgements}
We thank all reviewers for providing valuable feedbacks and suggestions for us to incorporate into the final version. The authors are supported by an NSERC Discovery Grant [RGPIN-2018-06806] This work is also supported in part by the Institute for Computing, Information and Cognitive Systems (ICICS) at the University of British Columbia. 

\bibliography{anthology,custom}
\bibliographystyle{acl_natbib}
\clearpage

\appendix

\section{Projected Attention Layer (PAL)}
\label{sec:pal-details}
Projected Attention Layer (PAL) proposed by \citet{stickland2019bert} as adaptor modules for the pretrained model. Similar to the design in the original work, the PAL layer runs "parallel" with the pretrained encoder layer where their respective output are added in a residual manner \citep{rebuffi2018efficient}, such that:
\begin{equation}
    \textrm{\textbf{h}}^{l+1} = \textrm{LN}\big(\textrm{\textbf{h}}^l + \textrm{SA}(\textrm{\textbf{h}}^l) + \textrm{PAL}(\textrm{\textbf{h}}^l)\big)
\end{equation}
where LN denotes LayerNorm, and SA is the self-attention layer in the pretrained encoder. In each PAL layer, the hidden size of pretrained layer is first reduced via linear projection and passed through its own self-attention layer before transformed back into the original hidden size.

\section{Experiment Settings}
\label{sec:exp-settings}

\subsection{Extraction Summarization}
\label{sec:summ-settings}
The dataset CNN/DM consists of news articles and multi-sentence highlights as summaries. In our work, we used the non-anonymized version processed by \citet{see-etal-2017-get} while following the standard dataset split that contains 287,226 training examples, 13,368 validation examples and 11,490 test examples \footnote{https://github.com/abisee/cnn-dailymail}. Following previous work \citep{xu-durrett-2019-neural, zhang-etal-2019-hibert, xu-etal-2020-discourse}, we create the NYT-50 dataset from the New York Times Annotated Corpus by removing the documents whose summaries are shorter than 50 words, and use the data split that consists of 137,778 training examples, 17,222 validation examples and 17,223 test examples. In both datasets, we use the same data pre-processing steps from previous work \citep{liu-lapata-2019-text, xu-etal-2020-discourse}, and obtain sentence-level oracle labels for extractive summarization by greedily select sentences that maximizes the ROUGE evaluation metric \citep{nallapati2017summarunner}. During training and inference, the documents are truncated to $512$ and $800$ tokens, respectively, for the CNN/DM and NYT-50 datasets.

During training, we use the ADAM optimizer \citep{kingma2015adam} ($\beta_1 = 0.9$, $\beta_2 = 0.999$) following the same learning rate scheduler used in \citep{liu-lapata-2019-text}. We train all our models for a total of $50,000$ steps where the validation loss is evaluated every $1,000$ steps for selecting the top-3 checkpoints. We perform all our experiments on a combination of NVIDIA GTX 1080 Ti and V100 under the single GPU setting, where the true batch size is set to $36$ with gradient accumulation per step is set to $9$ or $3$ for 1080 Ti and V100 respectively due to memory constraints. 

\subsection{Topic Segmentation}
\label{sec:topic-settings}
The dataset WikiSection \citep{arnold-etal-2019-sector} consists of Wikipedia documents with distinct section and subsection headings indicating topic boundaries. In our work, we use the largest English subset in city domain (\textit{en\_city}) consisting of 19.5k documents, and use the same $70$/$10$/$20$ (train/dev/test) split setting used by the authors\footnote{https://github.com/seastianarnold/WikiSection}. Similarly, Wiki-727 \citep{koshorek-etal-2018-text} consists of 727,000 open-domain documents from English Wikipedia, where we use the $80$/$10$/$10$ (train/dev/test) split setting used by the authors\footnote{https://github.com/koomri/text-segmentation}. During training and inference, the context length for the left and right windows are both set to $128$ tokens.

During training, we use the AdamW optimizer \citep{loshchilov2018decoupled} ($\beta_1 = 0.9$, $\beta_2 = 0.999$) following the same learning rate scheduler used in \citet{lukasik-etal-2020-text}. Due to the significant size difference between the two datasets, we trained the model on WikiSection for five epochs and on Wiki-727 for one epoch, where validation process is executed every $2,500$ steps to select the top-3 checkpoints. We perform all our experiments on a combination of NVIDIA GTX 1080 Ti and V100 under the single GPU setting, where the true batch size is set to $64$ with gradient accumulation per step is set to $4$ or $1$ for 1080 Ti and V100 respectively due to memory constraints.

\begin{figure*}[ht]
    \centering
    \includegraphics[width=0.8\linewidth]{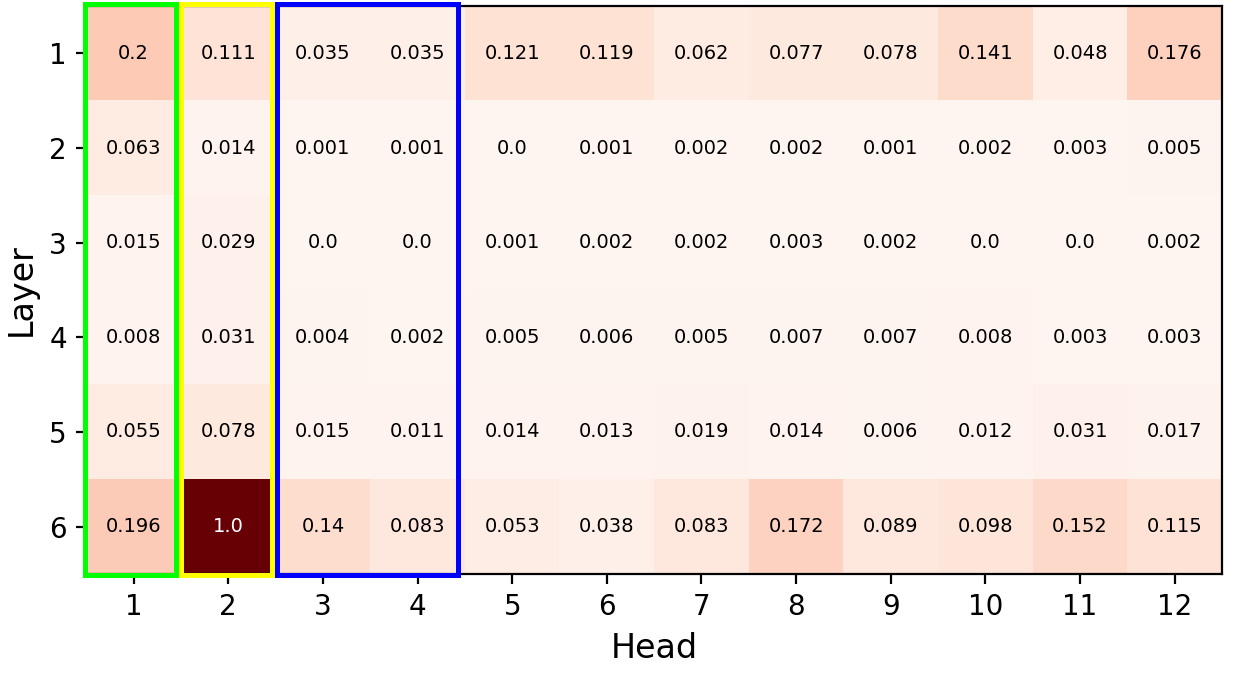}
    \caption{Importance heatmap for the 6-layer 12-head model. Head 1-4 are injected with the patterns, where the highlighted boxes represent Matching Token (Green), Intra Sentence (Olive) and Positional (Blue) (-1, +1).}
    \label{fig:distilled-importance}
\end{figure*}
\section{Head Importance of Pattern-Injected Models}

\subsection{Attention Heads}
\label{sec:distilled-importance}
Similarly, We also visualize the head importance score (\autoref{fig:distilled-importance}) using the 6-layer 12-head model on NYT-50, where the first four heads (index 1-4) of each layer are injected with our patterns (matching token, intra-sentence and positional, respectively). From this example, we can see that the heads with patterns injected
% applied
are considered to be more important across almost all layers, with the most important head being the intra-sentence head in the last layer. This fits our intuition since the output of the last layer is used as the sentence-representation for the classifier.

\subsection{Projected Attention Layer Heads}
\label{sec:pal-importance}
\begin{figure*}[ht]
     \centering
     \begin{subfigure}[b]{0.23\linewidth}
         \centering
         \includegraphics[width=\linewidth]{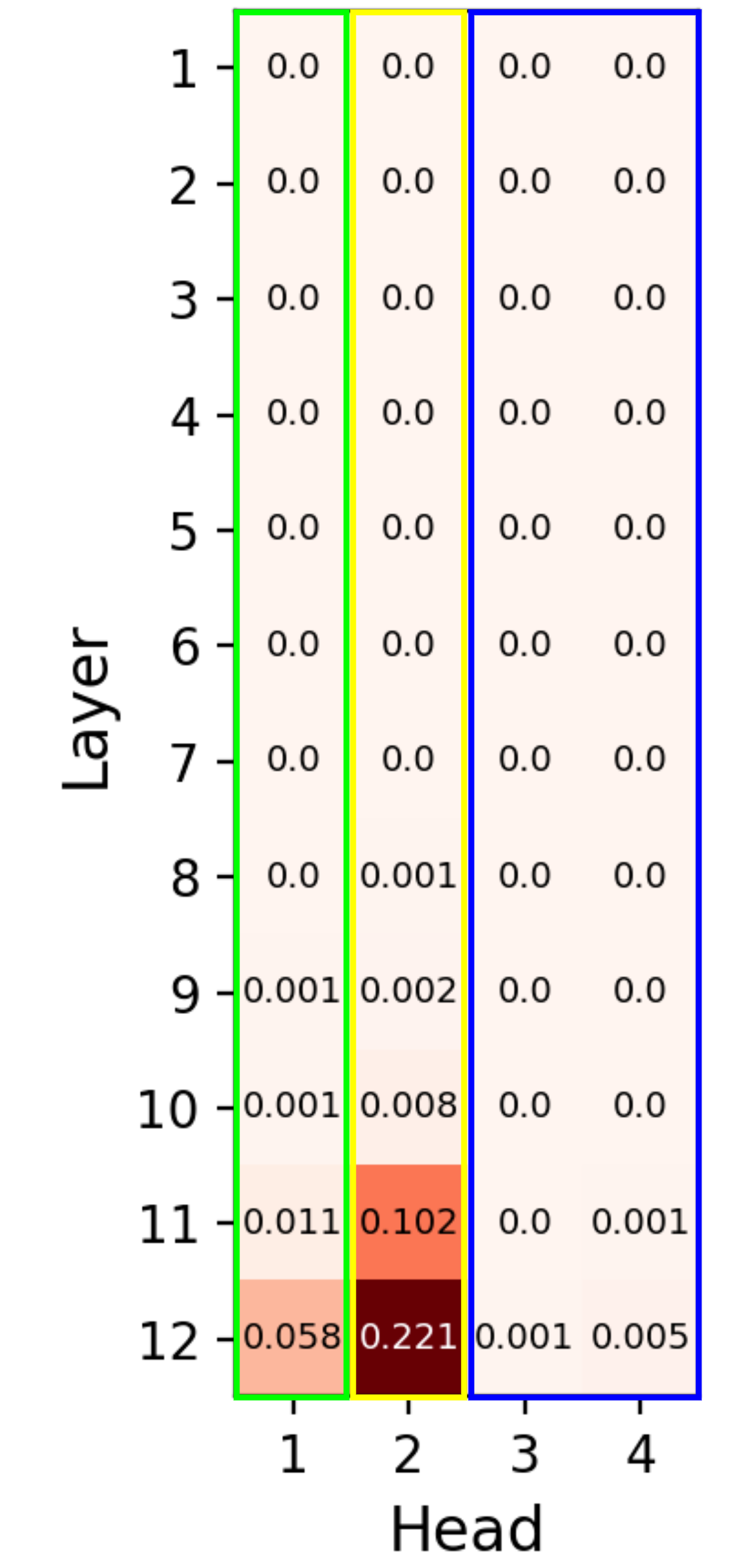}
         \caption{PAL with patterns}
         \label{fig:pal-patterns}
     \end{subfigure}
     \begin{subfigure}[b]{0.23\linewidth}
         \centering
         \includegraphics[width=\linewidth]{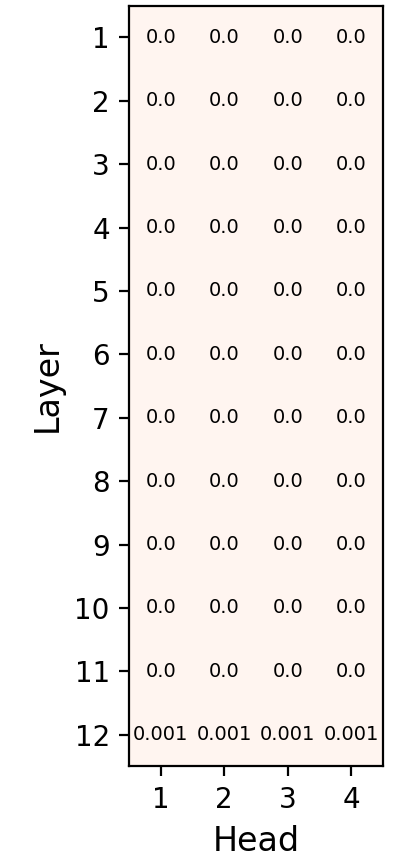}
         \caption{PAL without patterns}
         \label{fig:pal-vanilla}
     \end{subfigure}
     
     \caption{PAL head importance with (a) and without  (b) patterns injected, where the highlighted boxes represent Matching Token (Green), Intra Sentence (Olive) and Positional (Blue) (-1, +1)}
     \label{fig:pal-importance}
     \vspace{-1em}
\end{figure*}

We visualize the important scores of the PAL heads for BERTSum trained on CNN/DM (\autoref{fig:pal-importance}), where there are four heads added to each BERT layer via residual connection. \autoref{fig:pal-vanilla} shows the normalized importance score of the PAL heads without any patterns injected, where the model is opting to use almost entirely the representation from the BERT layers. In \autoref{fig:pal-patterns}, where each of the four PAL heads are injected with our patterns, we can see that importance score significantly increased from the score without the patterns injected, indicating that the features encoded in our patterns are indeed being utilized by the models in addition to the existing pretrained representations.

\section{Summarization with Trigram Blocking}
\vspace{-.7em}

\begin{table}[!ht]
    \centering
    \resizebox{\linewidth}{!}{
    \begin{tabular}{l|c|c|c|c|c|c}
    
    \multirow{2}{*}{Model}&\multicolumn{3}{c|}{CNN/DM}& \multicolumn{3}{c}{NYT-50}\\
    &R-1 & R-2 &R-L& R-1 & R-2 &R-L\\
    \hline
    \multicolumn{7}{c}{6 Layer 8 Head}\\
    \hline
    Transformer&41.07&18.41&37.45&45.13&26.05&37.52
\\
    +Patterns (4/8)&\textbf{41.93}&\textbf{19.04}&\textbf{38.37}&\textbf{46.36}&\textbf{27.03}&\textbf{38.58}\\
    \hline
    \multicolumn{7}{c}{6 Layer 12 Head}\\
    \hline
    Transformer&41.12&18.42&37.50&45.35&26.23&37.71\\
    +Patterns (4/12)&\textbf{42.01}&\textbf{19.12}&\textbf{38.44}&\textbf{46.09}&\textbf{26.84}&\textbf{38.35}\\
    \hline
    
    \multicolumn{7}{c}{6 Layer 12 Head (with BERT Embeddings)}\\
    \hline
    
    Transformer&41.38&18.65&37.79&45.35&26.20&37.66\\
    % DistillBERT&40.84&18.22&37.23&44.77&25.77&37.21\\
    % TinyBERT&42.16&\textbf{19.35}&38.61&\textbf{46.97}&\textbf{27.70}&\textbf{39.24}\\
    +Patterns (4/12)&\textbf{42.24}&\textbf{19.35}&\textbf{38.68}&\textbf{46.27}&\textbf{27.02}&\textbf{38.49}\\
    \end{tabular}}
    \caption{The results for the summarization experiments under three settings with Trigram Blocking applied.}
    \label{tab:kd-overall_triblck}
    \vspace{-1em}
\end{table}

\label{sec:trigram}
\subsection{Trigram Blocking}
In our experiments, we follow previous work \cite{paulus2018deep, liu-lapata-2019-text} in evaluating the models in two ways: with and without the trigram blocking. At inference time, the summary is usually formed by selecting sentences with the highest prediction scores. However, with the trigram blocking trick, sentences with overlapping trigram will not be selected. This trick has been shown to be an effective method to deal with redundancy on some dataset (e.g. CNN/DM), but may cause performance drop in others (e.g. Pubmed and arXiv). 

\begin{table}[!ht]
    \centering
    \resizebox{0.7\linewidth}{!}{
    \begin{tabular}{l|c|c|c|c|c|c}
    
    \multirow{2}{*}{Model}& \multicolumn{3}{c}{w Tri-block}\\
    & R-1 & R-2 &R-L\\
    \hline
    Transformer &41.07&18.41&37.45\\
    \quad + match (m)&+0.67&+0.54&+0.71\\
    \quad + intra (i)&+0.20&+0.12&+0.27\\
    \quad + pos (p) &-0.13&-0.13&-0.10\\
    \quad + m +i&+0.84&+0.57&+0.89\\
    \quad + m +p&+0.46&+0.38&+0.52\\
    \quad + i + p&+0.27&+0.20&+0.34\\
    \quad + all&\textbf{+0.86}&\textbf{+0.63}&\textbf{+0.92}\\
    \end{tabular}}
    \vspace{-.5em}
    \caption{Ablation study on the CNN/DM dataset (6-layer 8-head) with Trigram Blocking applied.}
    \vspace{-1em}
    \label{tab:kd-ablation_triblck}
\end{table}

\subsection{Pattern-Infused Sparse Transformers}
In \autoref{tab:kd-overall_triblck}, we show the trigram blocking results of the sparse transformer models on both summarization datasets, and \autoref{tab:kd-ablation_triblck} shows the trigram blocking results for pattern ablation experiment on the CNN/DM dataset. In line with \S\ref{sec:kd}, our pattern-infused models work better than all the other models on all of the settings on both dataset. As for the ablation study, we see a higher performance gain with the matching-token pattern when trigram blocking is applied, where the best performing model is still the one with all patterns applied.

% \cite{xiao-carenini-2020-systematically}.

\subsection{Guided Knowledge Injection into Pre-trained Models}

\autoref{tab:pal-results_triblck} shows that the results with trigram blocking. We find the performance gain from the patterns to be higher for CNN/DM and lower for NYT-50.

\begin{table}[ht]
    \centering
    \resizebox{\linewidth}{!}{
    \begin{tabular}{l|c|c|c|c|c|c}
    \multirow{2}{*}{Model} &  \multicolumn{3}{c|}{CNN/DM}& \multicolumn{3}{c}{NYT-50} \\
    &  R-1 & R-2 & R-L &  R-1 & R-2 & R-L \\
    \hline
    BERTSum  & 42.97 & 20.09 & 39.43&47.58 & 28.40 & 39.95\\
    + PAL & 42.96 & 20.07 & 39.41 & 47.78 & 28.56 & 40.15\\
    + PAL + Patterns & \textbf{43.07} & \textbf{20.12} & \textbf{39.50}& \textbf{48.25}&\textbf{29.10} & \textbf{40.70} \\
    \hline
    \end{tabular}}
    \caption{ROUGE F-scores of pretrained models with PAL when trigram blocking is applied.}
    \label{tab:pal-results_triblck}
    \vspace{-.5em}
\end{table}

\section{Pattern Ablation for Topic Segmentation}
\label{sec:topic-ablation}

\autoref{tab:topic-seg-ablation} shows that applying all 3 types of patterns leads to the highest performance gain in F-1 score. This is inline with the ablation results for extractive summarization.

\begin{table}[ht]
    \centering
    \resizebox{0.7\linewidth}{!}{
    \begin{tabular}{l|c|c|c|c|c|c}
    
    Model & P & R & F-1\\
    \hline
    Transformer & 0.698 & 0.647 & 0.671\\
    \quad + match (m)&+0.035&+0.064&+0.051\\
    \quad + intra (i)&+0.006&-0.006&+0.000\\
    \quad + pos (p) &+0.022&-0.022&-0.002\\
    \quad + m +i&+0.023&+\textbf{0.070}&+0.047\\
    \quad + m +p&+0.040&+0.059&+0.050\\
    \quad + i + p&+0.030&-0.030&-0.004\\
    \quad + all&\textbf{+0.046}&+0.064&\textbf{+0.056}\\
    \end{tabular}}
    \vspace{-.5em}
    \caption{Ablation study results on the WikiSection dataset with the 6-layer 8-head setting.}
    \label{tab:topic-seg-ablation}
    \vspace{-.5em}
\end{table}

\end{document}